\documentclass{article}

\usepackage{arxiv}
\usepackage{parskip}
\usepackage{wrapfig}

\usepackage{cite}
\usepackage{amsmath,amssymb,amsfonts}
\usepackage{graphicx}
\usepackage{textcomp}
\usepackage{tabularx}

\usepackage{cite}

\usepackage{bbold}
\usepackage{amsmath,amssymb,amsfonts}

\usepackage{textcomp}
\usepackage{subcaption}
\usepackage{verbatim}

\usepackage{algorithm}
\usepackage{algpseudocode}

\usepackage{hyperref}

\usepackage{amssymb}
\usepackage{amsmath}
\usepackage{multirow}
\usepackage{adjustbox}
\usepackage{booktabs}
\usepackage{tikz}
\usetikzlibrary{calc,shapes.geometric,arrows,positioning,intersections}
\usetikzlibrary{decorations, decorations.text,backgrounds}

\begin{document}

\title{Climate Change Impact on Agricultural Land Suitability:  A Machine Learning-Based\par Eurasia Case Study}

\author{Valeriy Shevchenko,
Daria Taniushkina,
Aleksander Lukashevich,
Aleksandr Bulkin,
Roland Grinis, \\
\textbf{Kirill Kovalev,
Veronika Narozhnaia,
Nazar Sotiriadi,
Alexander Krenke,
Yury Maximov}
}
\maketitle

\begin{abstract}
The United Nations has identified improving food security and reducing hunger as essential components of its sustainable development goals. As of 2021, approximately 828 million people worldwide are experiencing hunger and malnutrition, with numerous fatalities reported. Climate change significantly impacts agricultural land suitability, potentially leading to severe food shortages and subsequent social and political conflicts. To address this pressing issue, we have developed a machine learning-based approach to predict the risk of substantial land suitability degradation and changes in irrigation patterns. Our study focuses on Central Eurasia, a region burdened with economic and social challenges.

This study represents a pioneering effort in utilizing machine learning methods to assess the impact of climate change on agricultural land suitability under various carbon emissions scenarios. Through comprehensive feature importance analysis, we unveil specific climate and terrain characteristics that exert influence on land suitability. Our approach achieves remarkable accuracy, offering policymakers invaluable insights to facilitate informed decisions aimed at averting a humanitarian crisis, including strategies such as the provision of additional water and fertilizers. This research underscores the tremendous potential of machine learning in addressing global challenges, with a particular emphasis on mitigating hunger and malnutrition.

\end{abstract}


\maketitle


\section{Introduction}
\label{sec:introduction}
The impact of global climate change permeates various spheres of human activity, exerting a significant influence on global pandemics, food security, and political and social stability. With Earth's land surface facing alarming temperature increases and humidity decreases, the croplands and pastures that cover nearly 40\% of the planet's surface are under threat \cite{foley2005global}. Numerous studies suggest that global food demand may surge by approximately 110\% by 2050 \cite{godfray2010food,tilman2011global,shevchenko2023food,grabar2023longterm}. Additionally, the rising average temperature, snow-water equivalent, and carbon dioxide (CO${}_2$) concentration pose challenges to the optimal conditions required for crop growth \cite{gitz2016climate,huning2020global,porter2014food}.

This study examines the ramifications of climate change on agricultural and population sustainability under various Shared Socioeconomic Pathways (SSP) \cite{riahi2017shared} based on different climate conditions. The Coupled Model Intercomparison Project (CMIP) offers a coarse-grained assessment of critical climate indicators, including mean temperature, humidity, and atmospheric pressure. We utilize three robust CMIP6 models (CMCC-ESM2, CNRM-CM6-1, and MRI-ESM2-0) to evaluate three SSP scenarios \cite{peano2020cmcc,voldoire2019evaluation,SeijiYUKIMOTO20192019-051}: sustainable green energy (SSP1-2.6), business-as-usual (SSP2-4.5), and increased reliance on fossil fuels (SSP5-8.5). We also incorporate the Global Food Security-support Analysis Data at a Nominal 1 km (GFSAD) \cite{GFSSAD} to study cropland watering methods.

Our study addresses several key questions:
\begin{enumerate}
\item How does the distribution of croplands change due to climate change under different SSP scenarios?
\item Which areas face significant food security risks?
\item What are the primary factors influencing cropland suitability?
\end{enumerate}

We employ recurrent neural networks based on Long Short-Term Memory cells (LSTM, \cite{yu2019review}) to accurately classify agricultural land into four distinct classes: irrigation major, irrigation minor, rainfed, and minor cropland (non-cropland) based on prevailing climate conditions. Our findings suggest a significant expansion of land suitable for agriculture by the year 2050, particularly in the category of irrigation minor. This expansion has the potential to mitigate food-related risks in the region. Furthermore, our study predicts increased agricultural attractiveness in the northern part of the area while highlighting growing risks associated with currently utilized land. We extend our analysis to evaluate the evolution of land usage in Eastern Europe and Northern Asia by 2050.

By leveraging recurrent neural network models, we identify the critical factors affecting agricultural land suitability, including changes in irrigation patterns and land use patterns. Our study offers valuable insights into how land suitability will evolve over the next few decades and identifies the most influential factors shaping this evolution. Specifically, our research:
\begin{itemize}
\item Demonstrates the inherent dependencies between climate parameters and the risks associated with agricultural land use.
\item Investigates the evolution of agricultural land, pinpointing changes in irrigation patterns and suitability loss risks.
\item Identifies the key factors influencing land suitability for agricultural purposes.
\end{itemize}

Our findings hold particular relevance for policymakers and land use planners seeking to devise effective land management strategies that account for the complex interplay between climate, land suitability, and agricultural productivity.

Section \ref{sec:related_work} discusses relevant studies. We detail our approach, including analysis, datasets, and algorithmic considerations, in Section \ref{sec:Approach}. Section \ref{sec:evaluation} presents an empirical evaluation of several CMIP6 climate projections. We address the limitations of our study and potential areas for expansion in Section \ref{sec:constraints}. Finally, we conclude our study in Section \ref{sec:conclusion}.

\section{Related Work}
\label{sec:related_work}

Relevant studies fall into three categories: climate models, climate model analysis and forecasting, and neural networks.

\textbf{Climate Model Analysis:} Shoaib et al. \cite{shoaib2021quantifying} investigate the impact of different representative concentration pathways, namely RCP 4.5, RCP 6.0, and RCP 8.5, on crop yield in China using temperature and rainfall data. They employ a linear regression model to analyze crop yield based on the World Bank dataset \cite{worldbankdata}. Ramirez et al. \cite{muller2021exploring} analyze a total of 79 CMIP5 and CMIP6 projections of crop yield, their statistics, and dependence on pressure and temperature. The IPCC special report \cite{shukla2019ipcc} provides a general overview of climate projections regarding cropland suitability and outlines recommendations for policymakers.

\textbf{Climate Modeling:} The Climate Model Intercomparison Project phase 6 (CMIP6) \cite{eyring2016overview} provides experiments to describe future climate projections, representing the sixth phase of the project that began in 1995. Historical global reanalysis models such as ERA5 \cite{hersbach2020era5} and TerraClimate \cite{abatzoglou2018terraclimate} have also been developed. ERA5, with a spatial resolution of $31\times31$ kilometers, offers extensive temporal resolution support for 3-hour, 1-day, and 1-month resolutions and is continually refined in terms of physical models, data assimilation, and core dynamics. TerraClimate has a spatial resolution of $4\times4$ kilometers but is limited to monthly temporal resolution.

\textbf{Neural Networks Studies Related to Climate:} Tran et al. \cite{tran2021review} thoroughly review the application of artificial neural networks for temperature forecasting, while Dikshit et al. \cite{dikshit2022artificial} evaluate neural network architectures for drought prediction. Dharani and Chandrasekaran \cite{dharani2021review} provide a review of deep learning applications in crop prediction. They analyze approaches and models for different classification and regression problems in this field. Authors in \cite{diaconu2022understanding} employ ConvLSTM architecture for forecasting NDVI data. The paper \cite{yadav2021soil} studies classical machine learning models such as SVM and random forest for cropland fertility predictions. The study \cite{hounkpatin2022assessment} analyzes soil chemical compounds for Benin using classical machine learning.

\subsection{Approach}
\label{sec:Approach}

In this study, we propose a machine learning pipeline aimed at interconnecting overall climatic conditions and cropland classification by irrigation type. The pipeline utilizes climatic and morphological data as predictor variables to predict agricultural land conditions. This prediction relies on the variability of present combinations of land use, climate, and morphological variations. We assume that the link between the possibility of agricultural land use and climate conditions is invariant; thus, we can forecast future land use based on climate information if sufficient present variability is captured as a learning sample. However, this approach requires a significant amount of data, encompassing all possible states of climate and agriculture, including the absence of agriculture, related to a possible change in climate conditions for the area of interest.

We study the cropland types in Eastern Europe and Northern Asia to validate our proposed method. The study concludes that the territory of Russia, Eastern Europe, and Northern Asia provides sufficient diversity to generate all necessary examples for an agricultural land suitability study. To achieve this, we homogenize and align all data to the exact spatial and temporal resolution, perform feature design and data processing (Section \ref{sec:data}), and split the resulting dataset into training and test sets.

The machine learning models are then trained and evaluated in the second part of the pipeline. In the final part of the pipeline, we predict the distribution of cropland types based on different climate models, Shared Socioeconomic Pathways (SSP) scenarios, and analysis. The workflow also considers morphological data to constrain conditions where selected agricultural approaches are impossible.

\begin{figure}[!ht]
\centering
\includegraphics[width=\linewidth]{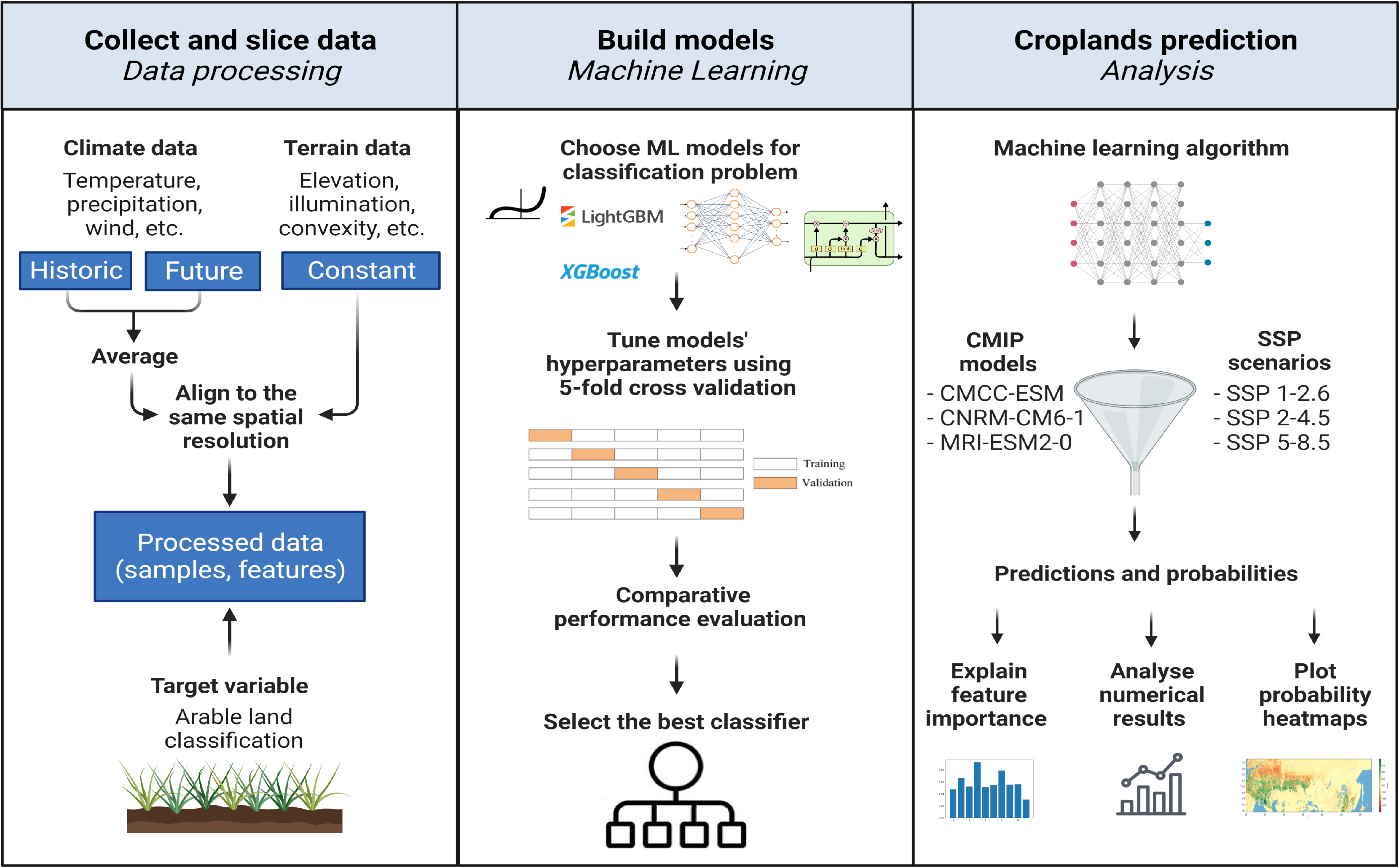}
\caption{The algorithmic workflow.}
\label{fig:workflow}
\end{figure}

In summary, our methodology adopts a three-part workflow that consists of data acquisition and processing, training machine learning models, and evaluating the results by predicting cropland distribution based on different climate models and SSP scenarios. The workflow provides stable results, utilizes historical data, and can simulate the distribution of cropland types in future climate projections. The workflow methodology is presented in Figure \ref{fig:workflow}.

\subsection{Data Acquisition and Processing}
\label{sec:data}

This subsection provides an overview of the data and preprocessing steps employed in this study.

Our study area spans from $19^\circ$E to $191^\circ$E and from $41^\circ$N to $78^\circ$N, covering diverse climate zones with varying degrees of agricultural development. We use the datasets listed in Table~\ref{tab:datasets} for reanalysis and modeling.

\begin{table*}
\caption{The dataset description.}
\label{tab:datasets}
    \centering
        \begin{tabular}{p{3.5cm}p{3cm}p{2cm}p{2.2cm}p{1.6cm}p{1.0cm}}
        \toprule
        Dataset Name & Bands used & Time \newline coverage & Spatial \newline resolution & Temporal \newline resolution & Type  \\
        
        \midrule
        SRTM Plus Elevation~\cite{elevation} &  Elevation & - & 30 arcsec & - & Model \\
        
        ERA5~\cite{hersbach2020era5} & Min and max temperature, precipitation & 1979–2021 & $\frac{1}{24}^\circ$ & Month & Model \\

        CMIP6~\cite{eyring2016overview} & 
        Monthly means of precipitation, min and max near-surface air temperatures & 1950–2100 &  $\frac{1}{2}^\circ-\frac{3}{2}^\circ$ &  Day & Model \\

        Five-class global cropland extent map~\cite{GFSSAD}   & Cropland watering method  & 2010 & $\frac{1}{20}^\circ$  & - & Model\\
        
        \bottomrule
    \end{tabular}

\end{table*}

To enhance accuracy, we employ the SRTM Plus Elevation data \cite{elevation} to obtain vertical elevation values. For historical climate data, we use ERA5 \cite{hersbach2020era5}, while future projections consider the SSP1-2.6, SSP2-4.5, and SSP5-8.5 simulations from CNRM-CM6-1 (Centre National de Recherches Météorologiques, France), CMCC-ESM2 (Centro Euro-Mediterraneo sui Cambiamenti Climatici, Italy), and MRI-ESM2-0 (The Meteorological Research Institute Earth System, Japan). These priority scenarios provide various socio-economic development paths, resulting in nine models under investigation. It's worth noting that ERA and CMIP-based analysis primarily focuses on mean value interpolation, often missing extreme fluctuations~\cite{morozov2023cmip}.

We use the Global Food Security Analysis Data (GFSAD) as a mask to identify croplands in our study area. The GFSAD is a 2010 nominal product dataset at a 1-km scale derived from remote-sensing data, secondary data, and field-plot data from 2009 to 2013 \cite{GFSSAD}. Initially, the croplands mask had five categories: irrigation major (class 1), irrigation minor (class 2), rainfed (class 3), rainfed minor fragments (class 4), and rainfed very minor fragments (class 5), with the first three classes accounting for approximately 75

The cropland mask assigns all objects to one of four categories related to cropland status, as shown in Table~\ref{tab:crop_class}. However, as the classes were highly imbalanced, we down-sample the majority class using random undersampling to ensure that all classes have a similar number of records in the dataset.

The terrain data we use includes elevation and ten other morphological variables calculated using four hierarchical levels equal to 3, 11, 33, and 47 kilometers. According to Turcotte \cite{turcotte_1997}, these four periodicity levels were chosen to capture the landscape on different spatial scales. We use spectral analysis to decompose the absolute relief height into individual frequencies and characterize the degree of expression of each frequency through its amplitude. Hierarchical levels correspond to the period. Fractal analysis also identifies model deviations and generates morphometric characteristics, as shown in Table~\ref{tab:Variables}. These features describe elevation gradient, surface shape, and shaded relief, influencing the risks of natural and anthropogenic activities while regulating heat and vapor distribution. 

 \begin{table}[!ht]
 \begin{center}
 {\caption{Initial cropland classes distributions.}\label{tab:crop_class}}
    \centering
    \begin{tabularx}{\columnwidth}{XXX}
    \toprule        
    \empty & Name & Amount  \\
    \midrule
    Class 0 & Minor cropland / & 11732309  \\
    \empty & non-cropland areas & \empty  \\
    Class 1 & Irrigation major &  146699 \\
    Class 2 & Irrigation minor & 586141 \\
    Class 3 & Rainfed & 1173203  \\
    \bottomrule
    \end{tabularx}
\end{center}
\end{table}



We collected monthly climate statistics from 2000 to 2010 as inputs and average them over ten years to reduce observation noise. This results in 12 values per climate feature, except for the SPI precipitation index, which generates a single annual value averaged as a single feature. We apply the same preprocessing procedure to future climate data for two time periods: 2020-2030 and 2040-2050.

The input model $X$ is represented as $x^\circ \times y^\circ$ and contains 121 climate features and 41 terrain features. The target variable is the cropland class label, which can be 0, 1, 2, or 3.

\begin{table*}[!ht]
\caption{Models features.}
\label{tab:Variables}
\begin{tabular}{p{0.7in}p{4.3in}p{1.0in}}
\toprule
\multicolumn{1}{c}{Title} & \multicolumn{1}{c}{Description}                                    & \multicolumn{1}{c}{Units}         \\ \midrule
\multicolumn{3}{c}{\textit{Climate data}}                                                                                                  \\ \midrule
tasmax   & frequency of days with extremely high temperatures (above the 95\% percentile)                                       & \hfil number of days \\
tasmin   & frequency of days with extremely low temperatures (below the 5\% percentile)                                         & \hfil number of days \\
t2m                                & monthly average air temperature at 2 meters                                 & \hfil °C                      \\
pr\_p95                            & frequency of days with extreme precipitation (above the 95\% percentile)    & \hfil number of days          \\
sfcWindmax                         & frequency of days with extreme wind speeds (above the 95\% percentile)      & \hfil number of days          \\
fy                                 & number of days per month with a Nesterov index of more than 4000°C          & \hfil number of days          \\
monT0ud                            & frequency of days with the air temperature passing through 0°C              & \hfil number of days          \\
monTstep6                          & frequency of days with daily temperature jumps of more than 6°C             & \hfil number of days          \\
12m\_SPI                           & standardized Precipitation Index                                            & \hfil -                       \\
tp                                 & monthly precipitation                                                       & \hfil mm/month                \\
snw                                & moisture content of the snow cover                                          & \hfil kg/m$^2$ \\ \midrule
\multicolumn{3}{c}{\textit{Terrain data}}                                                                                                  \\ \midrule
DEM\_1km                           & absolute altitude                                                           & \hfil m                       \\
morf\_1                            & slope                                                                   & \hfil °                       \\
morf\_2                            & aspect                                                                     & \hfil °                       \\
morf\_3  & shaded relief (cosine of the angle between the normal to the surface and the sun rays) from the east  & \hfil -              \\
morf\_4                            & profile convexity (profile in aspect/azimuth direction, vertical bulge) & \hfil -                       \\
morf\_5                            & plan convexity (horizontal bulge)                                                       & \hfil -                       \\
morf\_6  & longitudinal curvature (deviation of the surface from the normal to slope azimuth)                & \hfil -              \\
morf\_7  & cross section convexity (deviation of the surface from the normal to the slope)                      & \hfil 
 -              \\
morf\_8                            & minimum curvature (depicts saddle forms of relief)                                         & \hfil -                       \\
morf\_9                            & maximum curvature (depicts  hilled forms of relief)                                           & \hfil -                       \\
morf\_10 & shaded relief (cosine of the angle between the normal to the surface and the sun rays) from the south & \hfil -              \\ \bottomrule
\end{tabular}
\end{table*}

\subsection{Models}
The classification problem is addressed using supervised learning algorithms, specifically logistic regression, XGBoost \cite{Chen:2016:XST:2939672.2939785}, LightGBM \cite{ke2017lightgbm}, Multi-Layer Perceptron (MLP) \cite{haykin1994neural}, and Long Short-Term Memory (LSTM) \cite{hochreiter1997long}. For the LSTM, input features are reshaped into sequences of 12, corresponding to the number of months in a year. The historical data is divided into training and test subsets, allocating 75\% of the data for training and 25\% for testing, while both sets are normalized using the Min-max Scaler. Scaling parameters are determined based on the training set and applied to the test set and future data to avoid information leakage. A 5-fold cross-validation technique is employed to optimize the model's hyperparameters, while the test set is used for evaluation. The training data is shuffled to account for different climate zones in the study area. Additionally, for MLP and LSTM, 15\% of the test set is assigned to a validation set for tuning hyperparameters.

We denote $[P_{\mathcal{M}}^c]_{ij}$ as a machine learning (ML) model prediction for a Class $c$ for a pixel $ij$ from the climate projection/reanalysis model $\mathcal{M}$. We minimize the cross-entropy loss during training between the model output class probability and the current land use type from $\textup{GFSAD}$. For training the MLP and LSTM models, we use the Adam optimizer with a learning rate of $10^{-3}, ~\beta_1 = 0.9, ~ \beta = 0.999$ (for details, see \cite{kingma2014adam}). The inference procedure is detailed in Section \ref{sec:evaluation}.

As per the evaluation metrics presented in Table~\ref{tab:Metrics}, the LSTM model outperforms the other models, with an accuracy of 0.81 and a precision score of 0.79, indicating proficiency at correctly identifying positive cases. The F1 score of 0.73 suggests a good balance between precision and recall, while the average precision score of 0.82 implies that it ranks positive cases higher in the predicted list. Based on these results and previous research \cite{VANKLOMPENBURG2020105709,ZHONG2019430}, it can be concluded that the LSTM model is a strong candidate for the given classification task.

\begin{table}
\centering
\caption{Comparative performance assessment of ML Models for arable land prediction.}
\label{tab:Metrics}
\resizebox{%
      \ifdim\width>\columnwidth
        \columnwidth
      \else
        \width
      \fi
    }{!}{%
\begin{tabularx}{\columnwidth}{XXXXXX}
\toprule
Model &  Accuracy &
{\begin{tabular}[c]{@{}l@{}}Precision,\\ macro \\avg.\end{tabular}} &
{\begin{tabular}[c]{@{}l@{}}Recall,\\ macro \\avg.\end{tabular}} &
{\begin{tabular}[c]{@{}l@{}}F1-score,\\ macro\\ avg.\end{tabular}} &
{\begin{tabular}[c]{@{}l@{}}Average \\ Precision\\ score\end{tabular}} \\ \midrule
LR       & 0.66 & 0.63 & 0.52 & 0.54 & 0.58  \\
XGBoost  & 0.80 & 0.78  & 0.69 & 0.72 & 0.81 \\
LGBM & 0.78 & 0.76 & 0.66 & 0.69 & 0.77 \\
MLP      & 0.81  & 0.74 & 0.79 & 0.76 & 0.83 \\
LSTM     & 0.81 & 0.79 & 0.70 & 0.73 & 0.82\\ \bottomrule
\end{tabularx}%
}

\end{table}

\begin{table}[!ht]
\caption{Comparing performance of LSTM model for different classes.}
\label{tab:model_metrics_per_class}
\resizebox{%
      \ifdim\width>\columnwidth
        \columnwidth
      \else
        \width
      \fi
    }{!}{%
\begin{tabularx}{\columnwidth}{XXXXXX}
\toprule
Class &
  \multicolumn{1}{l}{Precision} &
  \multicolumn{1}{l}{Recall} &
  \multicolumn{1}{l}{F1-score} &
  \multicolumn{1}{l}{\begin{tabular}[c]{@{}l@{}}Average\\ Precision\\ score\end{tabular}} &
  \multicolumn{1}{l}{Support} \\ \midrule
0 &
  0.92 &
  0.87 &
  0.89 &
  0.96 &
  \multirow{4}{*}{\begin{tabular}[c]{@{}c@{}} 249306\\   31174\\ 124554\\ 249306\end{tabular}} \\
1 & 0.73 & 0.41 & 0.53 & 0.64 &  \\
2 & 0.75 & 0.66 & 0.70 & 0.79 &  \\
3 & 0.76 & 0.88 & 0.81 & 0.87 &  \\ \bottomrule
\end{tabularx}%
}

\end{table}

\subsection{Analysis}
\label{sec:Analysis}


We evaluate the LSTM performance for each class and provide the results in Table~\ref{tab:model_metrics_per_class}. The model's precision was consistently high across all classes, with class 0 exhibiting the highest precision of 0.92. The recall scores were relatively lower for classes 1 and 2, indicating that the model encountered difficulties in accurately identifying true positive instances. The F1-score, representing the harmonic mean of precision and recall, was highest for class 3, with a value of 0.81. The average precision score, measuring the area under the precision-recall curve, was high for all classes, signifying the model's effectiveness in ranking positive instances higher than negative ones.




We employed the permutation importance algorithm to enhance the interpretability of our black-box LSTM model. This widely accepted and reliable approach does not rely on any internal model parameters \cite{RJ-2009-013,PermImp}. This algorithm directly measures the importance of features by randomly shuffling all variables and observing the impact on the model score. In our case, the LSTM model was fitted on a test dataset $D$ of shape $(N, L, F)$, where $N$ represents the number of samples, $L$ the sequence length (in our case, 12 months), and $F$ the number of features (52 in our case). The precision score served as a reference score $s$. Each feature $j$ was randomly shuffled throughout the entire sequence, and the score $s_{k,j}$ of the model was computed for $k$ repetitions. The importance $I_j$ for each feature $f_j$ was defined as:


\begin{equation}
 I_j = s - \frac{1}{K}  \sum_{k=1}^{K}{s_{k,j}}
\end{equation}

This method of assessing LSTM permutation feature importance might influence the relationship between the feature and target variable. Consequently, the rate of decrease in the precision score for a specific feature can serve as an indicator of its importance to the model.



\begin{figure}[!ht]  \includegraphics[width=\textwidth]{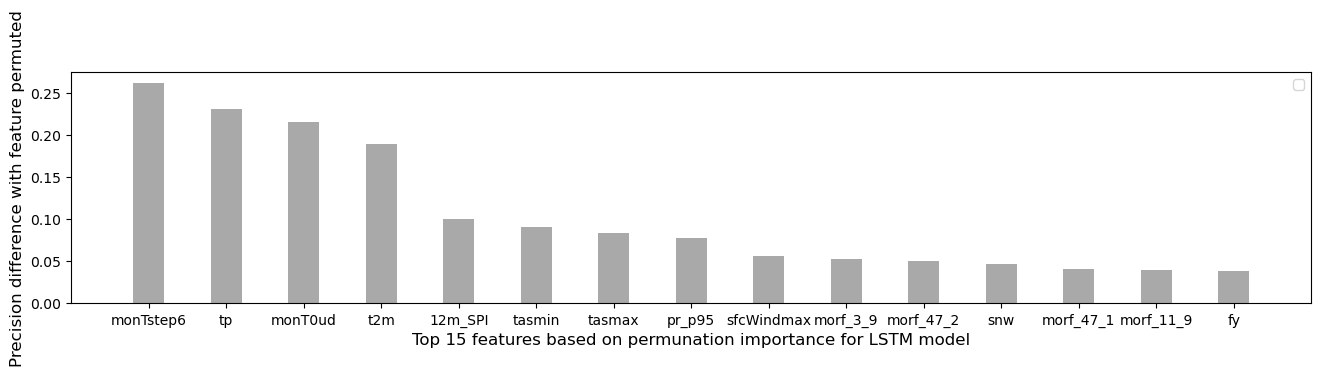}
  \caption{LSTM (permutation) feature importance.}
  \label{fig:feature_perm_imp}
 
\end{figure}

Our LSTM permutation feature importance analysis, depicted in Figure~\ref{fig:feature_perm_imp}, revealed that the frequency of days with daily temperature jumps exceeding 6°C and the monthly precipitation index were the most significant factors influencing the predictions. The frequency of days with daily temperature jumps greater than 6°C appears to be particularly important for the northern areas subject to agricultural expansion under the current climate change. Monthly precipitation is expected to directly affect the assignment of a class related to the irrigation strategy.  Moreover, among the terrain characteristics, the maximum curvature and aspect within 3 and 47 kilometers, respectively, were particularly important.

In summary, our approach provided valuable insights into the feature importance of an LSTM model, emphasizing the critical impact of certain weather variables and terrain characteristics on predictive outcomes.

\section{Performance Evaluation and Discussion}
\label{sec:evaluation}

The alteration of climatic conditions has the potential to impact various regions, transforming previously non-arable lands into croplands or vice versa. We developed an ML model based on climate data from 2000-2010, terrain data, and cropland masks from 2010 as a baseline. We aim to predict cropland types and quantities in Eastern Europe and Northern Asia by 2050, assuming no significant change in terrain characteristics. We considered three climatic models and three Shared Socioeconomic Pathways (SSP) scenarios and conducted nine experiments for the 2020-2030 and 2040-2050 periods. We assume that the applicability of climatic models may vary depending on the climatic zone selected, while SSP scenarios are generic. To prevent misinterpretation, we averaged the probabilities of each output by the climatic model post-training to obtain an ensemble prediction:

\begin{equation}
P^c_{ij} = \frac{\left[[P_{\textup{CNRM-CM6-1}}^c]_{ij}
+ [P_\textup{MRI-ESM2-0}^c]_{ij} + [P_\textup{CMCC-ESM2}^c]_{ij}\right]}{3}
\end{equation}

\subsection{Probabilistic Heatmaps}
\label{subsec:maps}
Regarding the SSP2-4.5 scenario, we aim to examine how cropland suitability varies across different locations based on the land use patterns of 2010, under various irrigation conditions, around the year 2050.

Let the LSTM model output for class $c$ at pixel $ij$ be denoted as $P^c_{ij}$, and let $P^c = \cup_{ij} P^c_{ij}$ be the heatmap for class $c$. First, we collect the output class probabilities $P^c_{ij}$ for each pixel from the LSTM model, obtaining probabilistic heatmaps $P^c$. Next, for each irrigation type, we subtract the heatmap $P^c$ from the 2010 usage for each class $\textup{GFSAD}^c$, with $\textup{GFSAD}_{ij}\in\{0, 1\}$, yielding $(P^c)_{\Delta}= P^c - \textup{GFSAD}^c$. 

A value close to $0$ indicates that the prediction aligns with the observed state of the land. While a value approaching $+1$ implies a likelihood that the current class $c$ will appear in the area, a value near $-1$ suggests the current class is at risk of being replaced. For instance, in the case of rainfed cropland, a difference close to $-1$ implies that the area may no longer be suitable for agricultural use.

\begin{figure}[!ht]
  \centering
             \hspace{-2mm} \subfloat[Major irrigated arable lands probability distribution]{\includegraphics[width=1.2\columnwidth]{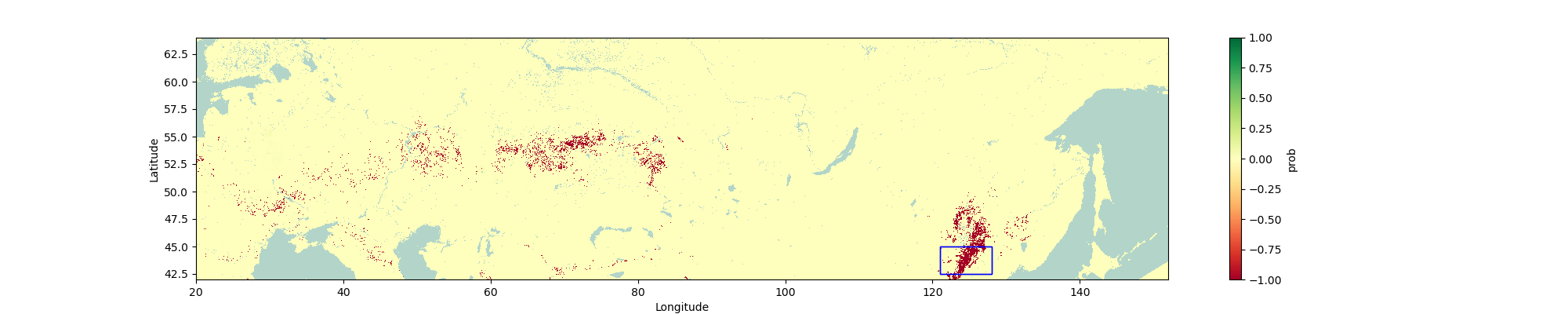}}
             
             \hspace{-2mm} \subfloat[Minor irrigated arable lands probability distribution]{\includegraphics[width=1.2\columnwidth]{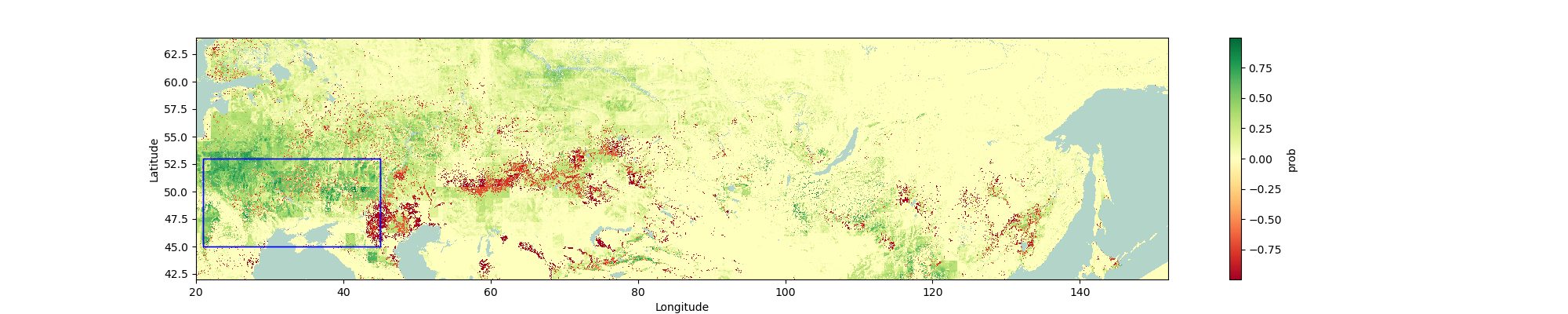}}
             
             \hspace{-2mm} \subfloat[Rainfed arable lands probability distribution]{\includegraphics[width=1.2\columnwidth]{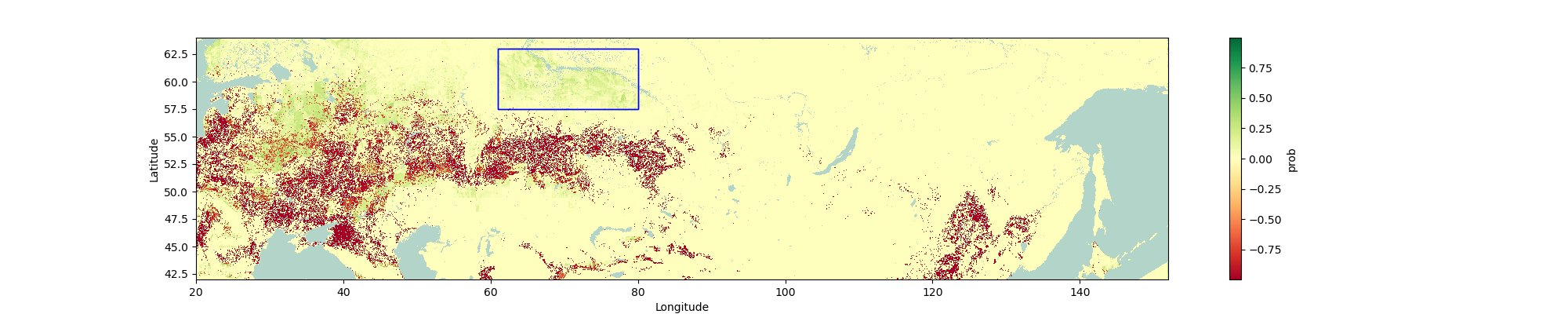}}

        \caption{Heatmaps of class probabilities from the LSTM model for arable lands classes for 2050 under SSP2-4.5 scenario. Blue rectangles stand for areas that are studies in Subsection \ref{subsec:local_feature}.}
        \label{fig:prob_heatmaps}
\end{figure}

According to the presented probabilistic heatmaps of $(P^c)_{\Delta}$ for each class in Figure~\ref{fig:prob_heatmaps}, our study suggests that various areas frequently utilized for agriculture are in danger, while the northern regions are becoming more appropriate for land usage without the requirement of irrigation. The results of our simulations also indicate a need to increase irrigation activity for the Eastern Europe zone due to the degradation of the rainfed arable land category. Therefore, it is essential to adopt a sustainable approach towards agriculture and land-use planning that takes into account the ecological carrying capacity of the local environment.

\begin{figure}[!ht]
\centering
\includegraphics[width=\linewidth]{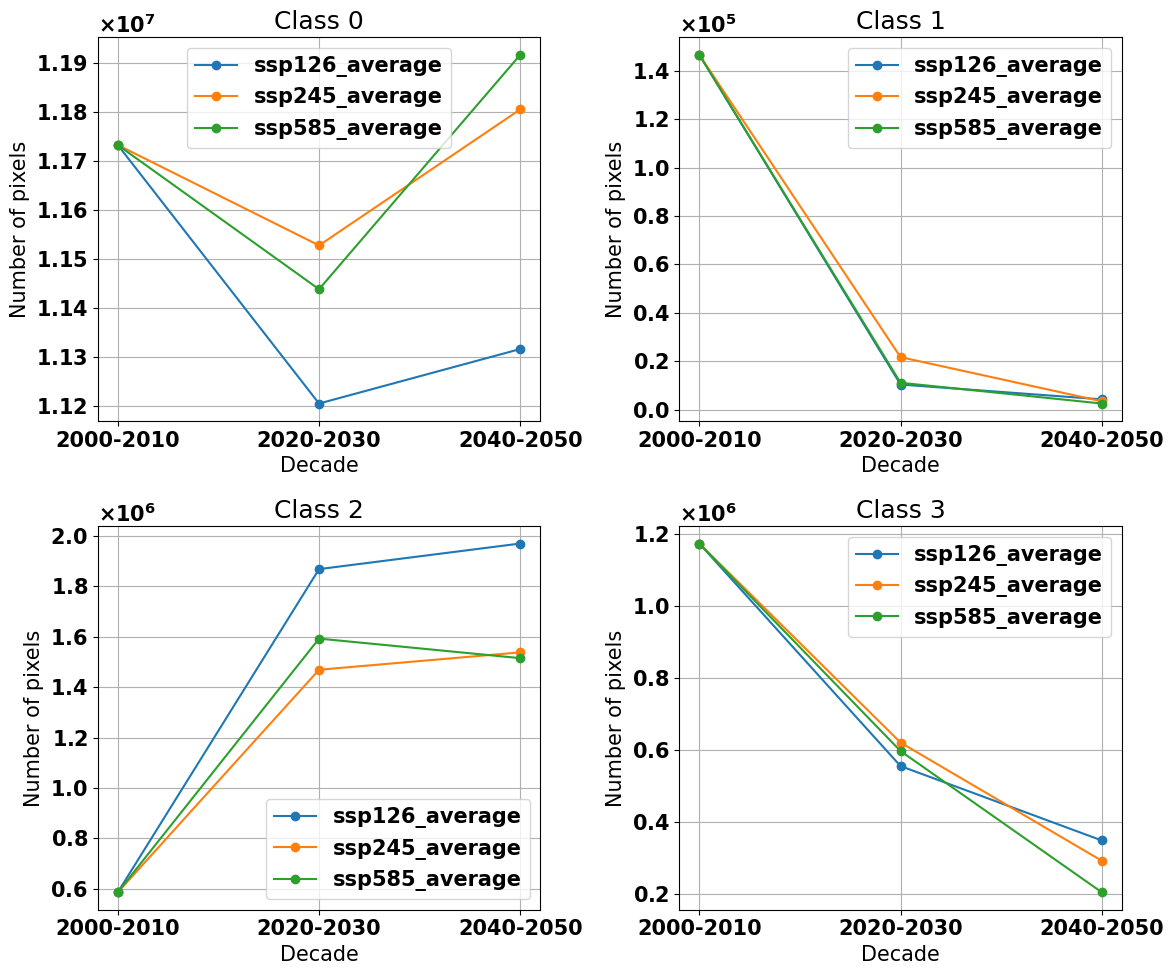}
  \caption{Pixel-wise cropland classification results forecasted by LSTM model under different SSP scenarios, with GFSAD cropland mask as the initial point}
  \label{fig:distribution_of_classes}
\end{figure}

Next, we examine how the number of pixels allocated to each class of croplands changes under various SSP scenarios until the year 2050 and present the distribution of cropland suitability in Figure~\ref{fig:distribution_of_classes}. Our analysis shows that, except for SSP1-2.6, there is a general trend toward increasing the suitability of lands for cropland use across all SSP scenarios. The SSP5-8.5 scenario displays the most positive trend, despite being environmentally unfriendly. Our findings are consistent with previous studies, including \cite{ramirez2017global}. The figures illustrate that the demand for irrigation (class 1) is anticipated to decrease, while the area of rainfed lands (class 3) will decline, and minor-irrigated (class 2) will expand. While this last class may indicate a further potential for development, more attention should be given to lands that transit from rainfed to irrigated, as they are already productive and are becoming more sensitive to water access. The overall trend is similar for all scenarios, but as greenhouse gas emissions increase, more lands are subject to transformation.

\subsection{Local Feature Attributes}
\label{subsec:local_feature}
In this section, we analyze the local specific feature importance for three regions: North-East China, Eastern Europe, and Northern Russia. These regions are likely to experience changes in their suitability for arable use, as discussed in Section \ref{subsec:maps}. North-East China is projected to lose suitability even with major irrigation (Class 1), while Eastern Europe is expected to require minor irrigation and lose its suitability as a rainfed cropland, moving from Class 3 to Class 2. On the other hand, Northern Russia is expected to become suitable for arable use under rainfed irrigation (Class 3). We have included the coordinates of the rectangles for these territories in Table \ref{tab:territories}.

\begin{table*}[!ht]
 \begin{center}
 {\caption{Regions with significant class change. Coordinates are given in latitude-longitude order. }\label{tab:territories}}
    \centering
    \begin{tabularx}{\textwidth}{p{0.8in}p{2.3in}p{0.8in}p{0.9in}p{0.9in}}
    \midrule      
    Zone & Subregions & Upper left corner & Lower right corner & Class change  \\
    \midrule
    Northern China & Inner Mongolia, Khingan, Jinlin, Liaoling & $45^\circ N, 121^\circ E$  & $42.5^\circ N, 128^\circ E$ & Arability loss  \\
    \midrule
    Eastern Europe & East Poland, Belarus, Ukraine, Slovakia, Hungary, Romania, Moldova, South Russia, West Kazakhstan & $53^\circ N, 21^\circ E$ &  $45^\circ N, 45^\circ E$ & Loss of rainfed cropland, need for minor irrigation \\
    \midrule
    Northern Russia & Khanty-Mansi District, Tomsk Region & $63^\circ N, 61^\circ E$ & $57.5^\circ N, 80^\circ E$ & Gain of suitable land, rainfed  \\
    \bottomrule
    \end{tabularx}
\end{center}
\end{table*}

To address local attribution, we employ the Integrated Gradients (IG) method \cite{sundararajan2017axiomatic}. According to the original paper, if $F(x)$ is a neural network with input $x=(x_1, \dots, x_n)$, where $x_i$ is an individual feature, then $A_F(x, x') = (a_1, \dots, a_n)$ is called an \emph{attribution} of the network prediction on $x$ relative to a baseline $x'$. Here, $a_i$ is a contribution of $x_i$ to the prediction $F(x)$. This method satisfies completeness, sensitivity, and implementation invariance.

Essentially, the IG attribution for a network $F$ is defined as $IG^F_i(x, x') = (x_i - x'_i) \times \int_{\alpha=0}^1 \frac{\partial F (x' + \alpha (x - x'))}{\partial x_i} d\alpha$. In other words, it is an integral of the gradient component corresponding to the feature $x_i$, between the baseline and the current input.

For a multi-class problem, we consider $F$ from above as a single logit of the class of interest. For example, if we aim to obtain attributions that impact either a shift to Class 3 (rainfed irrigation cropland) or away from it, we would choose $F$ to be the output of our model for this class.

\begin{figure}
    \centering
    \includegraphics[width=0.9\textwidth]{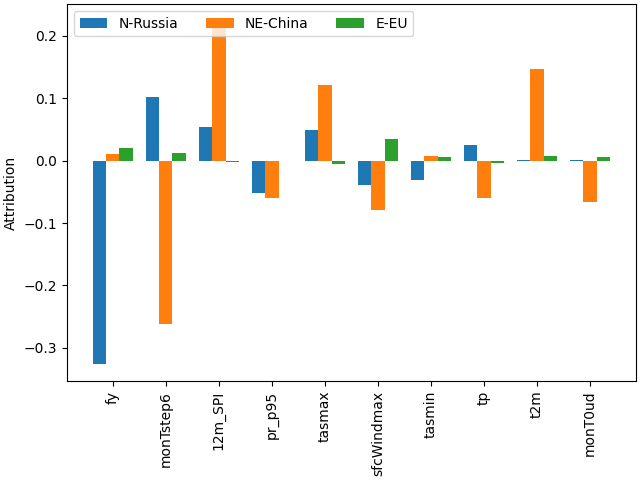}
    \caption{Top 10 local specific attributes obtained from IG, zones are defined in Table \ref{tab:territories}.}
    \label{fig:IG}
\end{figure}

We then analyze the territories above to interpret the LSTM model using the IG attribution method. We average the estimated attributions over the corresponding territories and present them in Figure~\ref{fig:IG}. We analyze impact of features for predictions for Classes 1, 2 and 3 for North-East China, Eastern Europe and Northern Russia, respectively. We chose historical climate data as a baseline, and as input, we used the 2050 climate state under the SSP2-4.5 (CNRM model) for these territories.

The results show that the Nesterov ignition index (fy) tends to have a negative influence on the suitability of cropland in Northern Russia. On the other hand, the attribution of the drought severity index (SPI) is positive for NE China as well as for Northern Russia. Moreover, the most significant negative impact on the loss of land suitability for China is made by the frequency of days between daily jumps in air temperature of more than 6 degrees (monTstep6). However, this feature turns out to be benevolent in Eastern Europe and Northern Russia, where it could increase land suitability with required irrigation.

\section{Constraints of this Study}
\label{sec:constraints}

The present study is delimited by the quality of the datasets utilized. GFSAD dataset has its limitations in terms of accuracy for cropland classes, finite spatial resolution, and the ERA5 and CMIP6 experiments, which are prone to overlooking crucial patterns and magnitudes of variables. Therefore, this study serves as a model that can demonstrate the qualitative behavior of cropland suitability under various scenarios but does not claim to be precise and accurate. Notice, that ERA5 and CMIP6 models are a good fit for the mean value analysis presented in the paper, while the impact of climate extremes is subjected to a separate study~\cite{morozov2023cmip,mozikov2023accessing,abramov2023advancing}. 

It is important to acknowledge that accurate crop mapping and monitoring remain a considerable challenge, and the chosen datasets present only a partial representation of reality. Specifically, the GFSAD dataset employed to define the land usage mask of the study is limited in its accuracy for crop classification, functioning better as a model rather than being entirely precise. Despite its limitations, this dataset enables the analysis of changes in cropland suitability by simulating land use classes.

Moreover, the ERA5 climate model has shown some limitations and biases in capturing specific climatic conditions, leading to a possible overestimation, underestimation, or misspecification of the climate variables. Therefore, the results produced should be considered as an approximation of possible scenarios of cropland suitability in the future and should not be seen as absolute or prescriptive. 

Lastly, the CMIP6 experiments used in this study, which provide an extensive range of future climate projections, have drawbacks, including the uncertainty and bias present in the model outputs. However, the scenarios created in this study can provide insights into potential future cropland suitability changes and give policymakers the basis to prepare and adjust their policies and decisions accordingly.

\section{Conclusion}
\label{sec:conclusion}

This study aimed to forecast the irrigation status of croplands under varying climatic conditions using different supervised learning techniques, namely logistic regression, XGBoost, LightGBM, Multi-Layer Perceptron (MLP), and Long Short-Term Memory (LSTM). Through an evaluation process, it was determined that LSTM was the most effective model for the classification task in terms of precision. As a result, the study proceeded to analyze three potential future climate projections (CNRM-CM6-1, CMCC-ESM2, MRI-ESM2-0) linked with SSP1-2.6, SSP2-4.5, and SSP5-8.5 scenarios until the year 2050 using the LSTM model.

Monthly average air temperature at 2 meters and monthly precipitation features were found to be critically important for classification. Climatic and morphological variations exhibited a strong relationship with land watering classes. Localized attribution analysis for regions in China and Eastern Europe, which are at risk of arability degradation, revealed that temperature increase played a significant role. Specifically, a temperature increase exceeding $6^\circ$ was found to be responsible for the degradation of crop lands in North-Eastern China.

Model improvement can be achieved by incorporating soil properties and indicators describing land water accessibility. The distribution of cropland classes was analyses on the map, averaging predictions of three future climate simulations. The results are presented separately for three SSP scenarios. The total amount of land suitable for agriculture is projected to increase, expanding further to the north. However, some currently exploited agricultural regions may require increased irrigation, posing potential risks.

Summing up, this study highlights the considerable potential for land development and the possibility of satisfying growing food demand. However, realizing these objectives necessitates intelligent land management and substantial investments. Failure to consider climate change may result in significant cropland shrinkage and subsequent deterioration of food security.
The repository \url{https://github.com/makboard/ArableLandSuitability} allows to replicate the analysis presented in the paper and gives the references to the input data.




\bibliographystyle{abbrv}
\bibliography{biblio}

\begin{thebibliography}{10}
\providecommand{\url}[1]{#1}
\csname url@samestyle\endcsname
\providecommand{\newblock}{\relax}
\providecommand{\bibinfo}[2]{#2}
\providecommand{\BIBentrySTDinterwordspacing}{\spaceskip=0pt\relax}
\providecommand{\BIBentryALTinterwordstretchfactor}{4}
\providecommand{\BIBentryALTinterwordspacing}{\spaceskip=\fontdimen2\font plus
\BIBentryALTinterwordstretchfactor\fontdimen3\font minus \fontdimen4\font\relax}
\providecommand{\BIBforeignlanguage}[2]{{%
\expandafter\ifx\csname l@#1\endcsname\relax
\typeout{** WARNING: IEEEtran.bst: No hyphenation pattern has been}%
\typeout{** loaded for the language `#1'. Using the pattern for}%
\typeout{** the default language instead.}%
\else
\language=\csname l@#1\endcsname
\fi
#2}}
\providecommand{\BIBdecl}{\relax}
\BIBdecl

\bibitem{foley2005global}
J.~A. Foley, R.~DeFries, G.~P. Asner, C.~Barford, G.~Bonan, S.~R. Carpenter, F.~S. Chapin, M.~T. Coe, G.~C. Daily, H.~K. Gibbs \emph{et~al.}, ``Global consequences of land use,'' \emph{science}, vol. 309, no. 5734, pp. 570--574, 2005.

\bibitem{godfray2010food}
H.~C.~J. Godfray, J.~R. Beddington, I.~R. Crute, L.~Haddad, D.~Lawrence, J.~F. Muir, J.~Pretty, S.~Robinson, S.~M. Thomas, and C.~Toulmin, ``Food security: the challenge of feeding 9 billion people,'' \emph{science}, vol. 327, no. 5967, pp. 812--818, 2010.

\bibitem{tilman2011global}
D.~Tilman, C.~Balzer, J.~Hill, and B.~L. Befort, ``Global food demand and the sustainable intensification of agriculture,'' \emph{Proceedings of the national academy of sciences}, vol. 108, no.~50, pp. 20\,260--20\,264, 2011.

\bibitem{shevchenko2023food}
V.~Shevchenko, A.~Lukashevich, A.~Bulkin, R.~Grinis, K.~Kovalev, V.~Narozhnaia, N.~Sotiriadi, A.~Krenke, and Y.~Maximov, ``Climate change and future food security: Predicting the extent of cropland gain or degradation,'' ArXiv:2310.03261.

\bibitem{grabar2023longterm}
V.~Grabar, A.~Marusov, Y.~Maximov, N.~Sotiriadi, A.~Bulkin, and A.~Zaytsev, ``Long term drought prediction using deep neural networks based on geospatial weather data,'' ArXiv:2309.06212, 2023.

\bibitem{gitz2016climate}
V.~Gitz, A.~Meybeck, L.~Lipper, C.~D. Young, and S.~Braatz, ``Climate change and food security: risks and responses,'' \emph{Food and Agriculture Organization of the United Nations (FAO) Report}, vol. 110, pp. 2--4, 2016.

\bibitem{huning2020global}
L.~S. Huning and A.~AghaKouchak, ``Global snow drought hot spots and characteristics,'' \emph{Proceedings of the National Academy of Sciences}, vol. 117, no.~33, pp. 19\,753--19\,759, 2020.

\bibitem{porter2014food}
J.~R. Porter, L.~Xie, A.~J. Challinor, K.~Cochrane, S.~M. Howden, M.~M. Iqbal, D.~B. Lobell, and M.~I. Travasso, \emph{Food security and food production systems}.\hskip 1em plus 0.5em minus 0.4em\relax Cambridge University Press, 01 2014, pp. 485--533.

\bibitem{riahi2017shared}
K.~Riahi, D.~P. Van~Vuuren, E.~Kriegler, J.~Edmonds, B.~C. O’neill, S.~Fujimori, N.~Bauer, K.~Calvin, R.~Dellink, O.~Fricko \emph{et~al.}, ``The shared socioeconomic pathways and their energy, land use, and greenhouse gas emissions implications: An overview,'' \emph{Global environmental change}, vol.~42, pp. 153--168, 2017.

\bibitem{peano2020cmcc}
D.~Peano, T.~Lovato, and S.~Materia, ``Cmcc cmcc-esm2 model output prepared for cmip6 ls3mip,'' \emph{Earth System Grid Federation: Italy}, 2020.

\bibitem{voldoire2019evaluation}
A.~Voldoire, D.~Saint-Martin, S.~S{\'e}n{\'e}si, B.~Decharme, A.~Alias, M.~Chevallier, J.~Colin, J.-F. Gu{\'e}r{\'e}my, M.~Michou, M.-P. Moine \emph{et~al.}, ``Evaluation of cmip6 deck experiments with cnrm-cm6-1,'' \emph{Journal of Advances in Modeling Earth Systems}, vol.~11, no.~7, pp. 2177--2213, 2019.

\bibitem{SeijiYUKIMOTO20192019-051}
S.~Yukimoto, H.~Kawai, T.~Koshiro, N.~Oshimo, K.~Yoshida, S.~Urakawa, H.~Tsujino, M.~Deushi, T.~Tanaka, M.~Hosaka, S.~Yabu, H.~Yoshimura, E.~Shindo, R.~Mizuta, A.~Obata, Y.~Adachi, and M.~Ishii, ``The meteorological research institute earth system model version 2.0, mri-esm2.0: Description and basic evaluation of the physical component,'' \emph{Journal of the Meteorological Society of Japan. Ser. II}, vol. advpub, pp. 2019--051, 2019.

\bibitem{GFSSAD}
P.~Teluguntla, P.~Thenkabail, M.~Xiong, Gumma, G.~Chandra, M.~Cristina, M.~Ozdogan, R.~Congalton, J.~Tilton, T.~Sankey, A.~Massey, Phalke, and K.~Yadav, \emph{Global Food Security Support Analysis Data at Nominal 1 km (GFSAD1km) Derived from Remote Sensing in Support of Food Security in the Twenty-First Century: Current Achievements and Future Possibilities, Chapter 6}.\hskip 1em plus 0.5em minus 0.4em\relax CRC Press, 11 2015, pp. 131--160, {ISBN:} 978-14-822-1795-7.

\bibitem{yu2019review}
Y.~Yu, X.~Si, C.~Hu, and J.~Zhang, ``A review of recurrent neural networks: Lstm cells and network architectures,'' \emph{Neural computation}, vol.~31, no.~7, pp. 1235--1270, 2019.

\bibitem{shoaib2021quantifying}
S.~A. Shoaib, M.~Z.~K. Khan, N.~Sultana, and T.~H. Mahmood, ``Quantifying uncertainty in food security modeling,'' \emph{Agriculture}, vol.~11, no.~1, p.~33, 2021.

\bibitem{worldbankdata}
WorldBank, ``World bank group, climate change knowledge portal.'' \url{https://climateknowledgeportal.worldbank.org/download-data}, 2023, accessed: 2023-02-27.

\bibitem{muller2021exploring}
C.~M{\"u}ller, J.~Franke, J.~J{\"a}germeyr, A.~C. Ruane, J.~Elliott, E.~Moyer, J.~Heinke, P.~D. Falloon, C.~Folberth, L.~Francois \emph{et~al.}, ``Exploring uncertainties in global crop yield projections in a large ensemble of crop models and cmip5 and cmip6 climate scenarios,'' \emph{Environmental Research Letters}, vol.~16, no.~3, p. 034040, 2021.

\bibitem{shukla2019ipcc}
P.~R. Shukla, J.~Skea, E.~Calvo~Buendia, V.~Masson-Delmotte, H.~O. P{\"o}rtner, D.~Roberts, P.~Zhai, R.~Slade, S.~Connors, R.~Van~Diemen \emph{et~al.}, ``Ipcc, 2019: Climate change and land: an ipcc special report on climate change, desertification, land degradation, sustainable land management, food security, and greenhouse gas fluxes in terrestrial ecosystems,'' 2019.

\bibitem{eyring2016overview}
V.~Eyring, S.~Bony, G.~A. Meehl, C.~A. Senior, B.~Stevens, R.~J. Stouffer, and K.~E. Taylor, ``Overview of the coupled model intercomparison project phase 6 ({CMIP6}) experimental design and organization,'' \emph{Geoscientific Model Development}, vol.~9, no.~5, pp. 1937--1958, 2016.

\bibitem{hersbach2020era5}
H.~Hersbach, B.~Bell, P.~Berrisford, S.~Hirahara, A.~Hor{\'a}nyi, J.~Mu{\~n}oz-Sabater, J.~Nicolas, C.~Peubey, R.~Radu, D.~Schepers \emph{et~al.}, ``The era5 global reanalysis,'' \emph{Quarterly Journal of the Royal Meteorological Society}, vol. 146, no. 730, pp. 1999--2049, 2020.

\bibitem{abatzoglou2018terraclimate}
J.~Abatzoglou, S.~Dobrowski, S.~Parks, and K.~Hegewisch, ``Terra{C}limate, a high-resolution global dataset of monthly climate and climatic water balance from 1958–2015,'' \emph{Scientific Data}, vol.~5, p. 170191, 01 2018.

\bibitem{tran2021review}
T.~T.~K. Tran, S.~M. Bateni, S.~J. Ki, and H.~Vosoughifar, ``A review of neural networks for air temperature forecasting,'' \emph{Water}, vol.~13, no.~9, p. 1294, 2021.

\bibitem{dikshit2022artificial}
A.~Dikshit, B.~Pradhan, and M.~Santosh, ``Artificial neural networks in drought prediction in the 21st century--a scientometric analysis,'' \emph{Applied Soft Computing}, vol. 114, p. 108080, 2022.

\bibitem{dharani2021review}
M.~Dharani, R.~Thamilselvan, P.~Natesan, P.~Kalaivaani, and S.~Santhoshkumar, ``Review on crop prediction using deep learning techniques,'' in \emph{Journal of Physics: Conference Series}, vol. 1767, no.~1.\hskip 1em plus 0.5em minus 0.4em\relax IOP Publishing, 2021, p. 012026.

\bibitem{diaconu2022understanding}
C.-A. Diaconu, S.~Saha, S.~G{\"u}nnemann, and X.~X. Zhu, ``Understanding the role of weather data for earth surface forecasting using a convlstm-based model,'' in \emph{Proceedings of the IEEE/CVF Conference on Computer Vision and Pattern Recognition}, 2022, pp. 1362--1371.

\bibitem{yadav2021soil}
J.~Yadav, S.~Chopra, and M.~Vijayalakshmi, ``Soil analysis and crop fertility prediction using machine learning,'' \emph{Machine Learning}, vol.~8, no.~03, 2021.

\bibitem{hounkpatin2022assessment}
K.~O. Hounkpatin, A.~Y. Bossa, Y.~Yira, M.~A. Igue, and B.~A. Sinsin, ``Assessment of the soil fertility status in benin (west africa)--digital soil mapping using machine learning,'' \emph{Geoderma Regional}, vol.~28, p. e00444, 2022.

\bibitem{elevation}
J.~Becker, D.~Sandwell, W.~Smith, J.~Braud, B.~Binder, J.~Depner, D.~Fabre, J.~Factor, S.~Ingalls, S.~H. Kim, R.~Ladner, and K.~Marks, ``Global bathymetry and elevation data at 30 arc seconds resolution: {SRTM30} plus,'' \emph{Marine Geodesy - MAR GEODESY}, vol.~32, pp. 355--371, 11 2009.

\bibitem{morozov2023cmip}
V.~Morozov, A.~Gabidullin, A.~Lukashevich, A.~Kurdukova, and Y.~Maximov, ``Cmip x-mos: Improving climate models with extreme model output statistics,'' In Review, 2023.

\bibitem{turcotte_1997}
D.~L. Turcotte, \emph{Fractals and Chaos in Geology and Geophysics}, 2nd~ed.\hskip 1em plus 0.5em minus 0.4em\relax Cambridge University Press, 1997.

\bibitem{Chen:2016:XST:2939672.2939785}
\BIBentryALTinterwordspacing
T.~Chen and C.~Guestrin, ``{XGBoost}: A scalable tree boosting system,'' in \emph{Proceedings of the 22nd ACM SIGKDD International Conference on Knowledge Discovery and Data Mining}, ser. KDD '16.\hskip 1em plus 0.5em minus 0.4em\relax New York, NY, USA: ACM, 2016, pp. 785--794. [Online]. Available: \url{http://doi.acm.org/10.1145/2939672.2939785}
\BIBentrySTDinterwordspacing

\bibitem{ke2017lightgbm}
G.~Ke, Q.~Meng, T.~Finley, T.~Wang, W.~Chen, W.~Ma, Q.~Ye, and T.-Y. Liu, ``Lightgbm: A highly efficient gradient boosting decision tree,'' \emph{Advances in neural information processing systems}, vol.~30, pp. 3146--3154, 2017.

\bibitem{haykin1994neural}
S.~Haykin, \emph{Neural networks: a comprehensive foundation}.\hskip 1em plus 0.5em minus 0.4em\relax Prentice Hall PTR, 1994.

\bibitem{hochreiter1997long}
S.~Hochreiter and J.~Schmidhuber, ``Long short-term memory,'' \emph{Neural computation}, vol.~9, no.~8, pp. 1735--1780, 1997.

\bibitem{kingma2014adam}
D.~P. Kingma and J.~Ba, ``Adam: A method for stochastic optimization,'' \emph{arXiv preprint arXiv:1412.6980}, 2014.

\bibitem{VANKLOMPENBURG2020105709}
\BIBentryALTinterwordspacing
T.~{van Klompenburg}, A.~Kassahun, and C.~Catal, ``Crop yield prediction using machine learning: A systematic literature review,'' \emph{Computers and Electronics in Agriculture}, vol. 177, p. 105709, 2020. [Online]. Available: \url{https://www.sciencedirect.com/science/article/pii/S0168169920302301}
\BIBentrySTDinterwordspacing

\bibitem{ZHONG2019430}
\BIBentryALTinterwordspacing
L.~Zhong, L.~Hu, and H.~Zhou, ``Deep learning based multi-temporal crop classification,'' \emph{Remote Sensing of Environment}, vol. 221, pp. 430--443, 2019. [Online]. Available: \url{https://www.sciencedirect.com/science/article/pii/S0034425718305418}
\BIBentrySTDinterwordspacing

\bibitem{RJ-2009-013}
\BIBentryALTinterwordspacing
C.~Strobl, T.~Hothorn, and A.~Zeileis, ``{Party on!}'' \emph{{The R Journal}}, vol.~1, no.~2, pp. 14--17, 2009. [Online]. Available: \url{https://doi.org/10.32614/RJ-2009-013}
\BIBentrySTDinterwordspacing

\bibitem{PermImp}
A.~Altmann, L.~Tolosi, O.~Sander, and T.~Lengauer, ``Permutation importance: A corrected feature importance measure,'' \emph{Bioinformatics (Oxford, England)}, vol.~26, pp. 1340--7, 04 2010.

\bibitem{ramirez2017global}
N.~Y. Ramirez-Cabral, L.~Kumar, and F.~Shabani, ``Global alterations in areas of suitability for maize production from climate change and using a mechanistic species distribution model ({CLIMEX}),'' \emph{Scientific reports}, vol.~7, no.~1, pp. 1--13, 2017.

\bibitem{sundararajan2017axiomatic}
M.~Sundararajan, A.~Taly, and Q.~Yan, ``Axiomatic attribution for deep networks,'' in \emph{International conference on machine learning}.\hskip 1em plus 0.5em minus 0.4em\relax PMLR, 2017, pp. 3319--3328.

\bibitem{mozikov2023accessing}
M.~Mozikov, I.~Makarov, A.~Bulkin, D.~Taniushkina, R.~Grinis, and Y.~Maximov, ``Accessing convective hazards frequency shift under climate change with physics-informed machine learning,'' ArXiv:2310.03180, 2023.

\bibitem{abramov2023advancing}
D.~Abramov, L.~Kurochkina, Y.~Melkozerova, V.~Pyzh, E.~Ponomarev, R.~Grinis, and Y.~Maximov, ``Advancing hydrological simulation in eastern europe, central, and northern asia: a comprehensive database and a deep learning approach,'' In Review, 2023.

\end{thebibliography}

\end{document}